\documentclass[preprint,12pt]{elsarticle}

\usepackage{amssymb}
\usepackage{subfig}
\usepackage{booktabs}
\usepackage{multirow}
\usepackage{multicol}
\usepackage{graphicx}
\usepackage{xcolor}
\usepackage{soul}
\usepackage{amsmath}

\usepackage{algpseudocode}
\usepackage{algorithm2e}
\usepackage{bbold}

\journal{Information sciences}

\begin{document}

\begin{frontmatter}



\title{Edge Conditional Node Update Graph Neural Network for Multi-variate Time Series Anomaly Detection}


\author[1]{Hayoung Jo}
\author[1]{Seong-Whan Lee}

\affiliation[1]{organization={Department of Artificial Intelligence, Korea University},
    addressline={Anam-dong, Seongbuk-ku}, 
    city={Seoul},
    postcode={02841}, 
    country={Republic of Korea}}

\begin{abstract}
With the rapid advancement in cyber-physical systems, the increasing number of sensors has significantly complicated manual monitoring of system states. Consequently, graph-based time-series anomaly detection methods have gained attention due to their ability to explicitly represent relationships between sensors. However, these methods often apply a uniform source node representation across all connected target nodes, even when updating different target node representations. Moreover, the graph attention mechanism, commonly used to infer unknown graph structures, could constrain the diversity of source node representations. In this paper, we introduce the Edge Conditional Node-update Graph Neural Network (ECNU-GNN). Our model, equipped with an edge conditional node update module, dynamically transforms source node representations based on connected edges to represent target nodes aptly. We validate performance on three real-world datasets: SWaT, WADI, and PSM. Our model demonstrates 5.4\%, 12.4\%, and 6.0\% higher performance, respectively, compared to best F1 baseline models.

\end{abstract}



\end{frontmatter}


\section{Introduction}
\label{intro}
Anomaly detection, which identifies deviations from established norms, plays a critical role in various areas such as industrial operations \citep{wu2021graph,jin2023varying}, intrusion detection \citep{tang2021detection, marteau2021random, basati2022pdae}, and fraud prevention \citep{guo2019detecting, jin2021intrusion, liu2023slafusion}. In response to the challenges posed by the proliferation of sensors in cyber-physical systems such as industrial systems, vehicles, and data centers, there is an increasing reliance on multivariate time-series anomaly detection.
Machine learning techniques, have been already applied successfully in various complex tasks such as face recognition \citep{weyrauch2004component}, video shot detection \citep{lee2001automatic}, action recognition \citep{ahmad2006human}, and pose estimation \citep{shakhnarovich2003fast}, have also found their application in multivariate time-series anomaly detection. However, these data have the complex interrelationships among numerous sensors with high levels of non-linearity. Traditional anomaly detection models, typically focusing on individual sensor data or employing modeling techniques with simple function, are inadequate for handling this complexity. Moreover, these conventional methods often lack scalability, leading to increased computational time as the number of sensors grows.

The advancement of deep learning has led to the proposal of various models like autoencoder-based models \citep{audibert2020usad, tayeh2022attention, zhou2022contrastive, zhang2023probabilistic} and generative adversarial networks (GAN) \citep{li2019mad, du2021gan, zhang2023stad, xu2022tgan} for learning diverse and complex normal distributions. Transformers, known for their success in natural language processing, have also been utilized due to their proficiency in modeling time dependencies in time-series data \citep{zhang2021unsupervised, chen2021learning, tuli2022tranad, wang2023disentangled}. Despite these advancements, these methods often provide meaningful insights only when the anomalies and detection sensors are the same. For instance, if an attacked sensor appears to function normally while another sensor shows abnormalities, pinpointing the compromised sensor becomes a challenge.

Graph-based methods have become popular in addressing this limitation, as they illustrate the relationships between sensors using a graph structure, offering additional insights in the event of anomalies. This enhanced understanding of sensor interconnections not only provides a deeper analysis of anomaly events but also accelerates the process of identifying causes and implementing corrective measures. However, the primary challenge with a graph-based approach is that the connections between sensors necessary for forming a graph structure are typically unknown. Therefore, learning the graph structure and detecting anomalies must occur simultaneously. 
Many previous studies have addressed this problem using a graph attention mechanism. In approaches such as \citep{zhao2020multivariate, guan2022gtad, zhou2022hybrid}, spatial relationships are implicitly learned through a graph attention mechanism, which operates under the assumption that all nodes are interconnected. In \citep{deng2021graph, ding2023mst}, node embedding vectors that represent node characteristics are introduced. These embedding vectors are used to extract graph structures based on the similarity of the vectors. During this process, the node embedding vectors are utilized to calculate the attention scores.

Previous graph-based methods apply the same source node representation to all connected target nodes, even when updating different target node representations. This approach is effective when source node representations are suitable for all connected targets, but this is often challenging in complex systems. It places a significant burden on the encoder to find a universally applicable source node representation. Additionally, using graph attention can act as an implicit constraint, forcing the source node representation to be similar.

To address these challenges, in this study, we propose Edge Conditional Node Update Graph Neural Network (ECNU-GNN), a novel graph neural network that dynamically updates target node representations. This is achieved by transforming source node representations into forms that are specifically tailored to each target node, guided by edge-specific conditions. This method learns the graph structure without relying on graph attention.
Our model comprises four main components: \textbf{node condition embedding}, \textbf{graph structure extraction}, \textbf{condition-based prediction}, and \textbf{anomaly detection}. \textbf{Node condition embedding} utilizes node embedding vectors as conditions to capture the characteristics of sensors. These embedding vectors are used to construct the graph structure, modify source representations, and read out target node representations. \textbf{Graph structure extraction} constructs the graph structure using the relationships among all pairs of node embedding vectors. \textbf{Condition-based prediction} updates the target node representations using different source node representations based on edges and predicts the next time step values from the updated target node representations. From the predicted values, \textbf{anomaly detection} judges whether an anomaly is detected at that time step.

To summarize, the main contributions of our work are:

\begin{enumerate}
\itemsep=0pt
\item We propose a state-of-the-art edge conditional node update graph neural network for time-series anomaly detection, featuring an edge conditional node update module. This module transforms source node representations based on edges, ensuring more adequate source node representation for updating the target node representations.
\item Our model constructs the graph structure without graph attention. Graph attention may act as an implicit regularization, leading to the encoder’s source node representations becoming more similar. This effect limits the variety of source node representations.
\item Our model excels in anomaly detection, outperforming baseline methods in tests on three real-world datasets. Our model features an innovative edge conditional node update module to effectively adjust source node representations to enhance target node updates and boost anomaly detection performance even in complex datasets. 

\end{enumerate}

The remainder of this paper is organized as follows. Section \ref{sec_2} describes the related work. The details of the proposed model are presented in Section \ref{sec_3}. A performance evaluation and other experimental results are presented in Section \ref{sec_4}. Finally, Section \ref{sec_5} concludes the paper.

\section{Related Work}
\label{sec_2}

\subsection{Graph Neural Networks}
Recently, graph neural networks (GNN) \citep{defferrard2016convolutional} have become a successful method for modeling intricate patterns in data containing graph structures, such as social networks and medical data. A graph comprises a collection of nodes linked by edges that signify the connections between these nodes. GNNs are particularly adept at modeling complex and malleable data structures, surpassing the capabilities of conventional deep neural networks.

Typically, GNNs assume that node representations are constructed based on the influences of neighboring nodes in a graph. Graph convolutional networks (GCN) \citep{kipf2016semi} aggregate the feature representations of nodes by directly considering the feature representations of neighbors.

The attention mechanism, which is a crucial component in numerous deep learning applications, is widely used in sequential models to efficiently capture relevant information. When applied to graphs, this concept gives rise to graph attention networks (GAT) \citep{velivckovic2017graph} that introduce a weighted aggregation of node representations during the aggregation step in GCN. In GAT, nodes are assigned varying weights through training, prioritizing more important nodes with higher weights and assigning lower weights to less significant ones.

\subsection{Time series anomaly detection}
Time-series anomaly detection has been the subject of extensive research for slightly some time, resulting in several proposed methodologies. Classical approaches are characterized by their simplicity and are ill-equipped to handle increasing system complexity. As deep learning has demonstrated exceptional performance in diverse tasks and various domains, numerous methods have emerged to address the challenges of time-series anomaly detection using deep learning techniques.

A fundamental structure for anomaly detection using deep learning is the autoencoder, which trains a network to reconstruct the input data of normal behavior and detects anomalies based on the difference between the input and the reconstructions. 
USAD \citep{audibert2020usad} is an autoencoder-based model equipped with two decoders. These two decoders are trained adversarially, with one decoder focused on producing accurate reconstructions and the other focused on producing reconstructions that differ significantly from the input data. This dual-decoder setup is designed to effectively detect anomalies that exhibited small deviations from normal data.
ACLAE-DT \citep{tayeh2022attention} employs an attention-augmented ConvLSTM and a dynamic thresholding mechanism in its reconstruction-based framework. In particular, it constructs as input an inter-correlation feature map that represents the correlations between all pairs of input data and entity embedding vectors, thereby capturing the temporal and contextual dependencies within the data.
CAE-AD \citep{zhou2022contrastive} introduces two contrasting methods: contextual contrasting and instance contrasting. The contextual contrasting method enables the attention-based projection layer to embed shared information between adjacent time steps. The instance contrasting method allows the encoder to capture shared information within a specific time step, utilizing data augmentation in both the time and frequency domains.
PAMFE \citep{zhang2023probabilistic} reconstructs clean inputs from noisy ones by using multiple parallel dilated convolutions with different dilation factors. This approach, combined with a feature fusion module, allows the model to capture features at different scales and increases its robustness to noise.

Generative Adversarial Networks (GAN)\citep{goodfellow2014generative} consist of a generator that generates data and discriminator that distinguishes between real and generated data. Methods using GANs typically train a generator and discriminator on normal time-series data, and compute anomaly scores based on the outputs of the generator or discriminator. 
MAD-GAN \citep{li2019mad} introduces inverse mapping to traditional GANs with LSTM-based generators and discriminators. It uses inverse transformations to compute anomaly scores using both the reconstruction loss and discriminator results. MAD-GAN aims to capture the hidden relationships and interactions that exist between sensors by considering the entire set of sensors simultaneously. 
FGANomaly \citep{du2021gan} generates pseudo-labels indicating whether the input is normal, based on the deviation between the original input and its reconstruction. This model is adversarially trained using data labeled 'normal' in the pseudo-labels, which increases its robustness to contaminated data.
STAD-GAN \citep{zhang2023stad} is designed within a teacher-student framework, combining a GAN structure with a classifier to overcome noise vulnerability and poor performance on anomaly-containing training data. Initially, the teacher model generates a refined dataset with pseudo-labels through unsupervised training. The student model then trains with this dataset, enhancing pseudo-label accuracy. This iterative process improves the classifier's anomaly detection capabilities.
TGAN-AD \citep{xu2022tgan} is a GAN model that includes a transformer-based generator and a discriminator designed to capture contextual information. The generator is trained to identify the latent vector with the highest covariation between the input and generated data, ensuring that the generated data closely resembles the original.

Recent research has extended the success of transformer models \citep{vaswani2017attention} in natural language processing to multivariate anomaly detection by adapting them to the time-series domain. 
TransAnomaly \citep{zhang2021unsupervised} integrates the transformer architecture with a variational autoencoder to tackle the issue of future dependency in sequential data modeling. This approach leads to a model that not only captures future dependencies more effectively but also offers reduced computational complexity, improved parallel processing, and enhanced explainability.
GTA \citep{chen2021learning} extracts graph structures through a learned network and encodes spatial features using GCN. Additionally, it employs transformer models to capture both global and local temporal features. This integrated approach facilitates anomaly detection by considering spatial and temporal dependencies effectively.
TranAD \citep{tuli2022tranad} is a transformer-based model using two decoders. The first decoder performs reconstruction, whereas the second decoder uses the difference between the results of the first decoder and the input to perform additional reconstruction. Both results contributed to the adversarial training process.
D$^{3}$TN \citep{wang2023disentangled} predicts the next time step value using a GRU-based predictor that processes features comprising global spatial features generated by a disentangled convolution layer, local spatial features, and temporal features derived from a self-attention-based layer.

Graph neural networks offer the advantage of explicitly defining the relationships between nodes. Leveraging this capability, they utilize signals from each sensor as node features to learn relationships between sensors, thereby aiding in anomaly detection.
MTAD-GAT \citep{zhao2020multivariate} is a graph attention-based model that constructs a graph structure without prior knowledge. It uses a fully connected graph across both spatial and temporal dimensions. To capture the unique characteristics of these dimensions, MTAD-GAT employs two distinct GAT models: one for spatial features and the other for temporal features. 
GDN \citep{deng2021graph} introduces node-embedding vectors to obtain a graph structure based on the similarity of the embedding vectors. Unlike conventional GATs, which rely solely on node features, a GDN concatenates embedding vectors with node features to compute the attention weight. 
GTAD \citep{guan2022gtad} extracts temporal features using a temporal convolution network and spatial features through a graph attention network. Utilizing these features, the model is trained for data reconstruction and prediction. Anomaly detection is then performed by analyzing both the reconstruction and prediction errors.
HAD-MDGAT \citep{zhou2022hybrid}, similar to MTAD-GAT \citep{zhao2020multivariate}, utilizes two GAT models to extract both spatial and temporal features. Leveraging these features, the model employs a GAN for input reconstruction and an MLP for data prediction.
MST-GAT \citep{ding2023mst} captures multimodal correlations by employing both inter-modal and intra-modal attention mechanisms, which leverage modality-specific information. Additionally, it utilizes multi-head attention that operates independently of modality information.
PGRGAT \citep{wu2023physics} employs prior information of sensor relationships to learn graph structure at an edge predictor. This graph structure is then used to encode features via the gated current graph attention unit network, which is a graph-based GRU module.

\section{Method}
\label{sec_3}
\subsection{Problem Statement}
The dataset consists of multivariate time-series data collected from $N$ sensors over a period of time $T$, $\boldsymbol{s}=[\boldsymbol{s}^{1},\boldsymbol{s}^{2},\dots,\boldsymbol{s}^{T}] \in \mathbb{R}^{N \times T}$.
Each column $\boldsymbol{s}^{t}$ at time $t$ corresponds to a vector representing measurements from $N$ sensors, denoted as $\boldsymbol{s}^{t} =[{s}^{t}_{1}, \dots, {s}^{t}_{N}] \in \mathbb{R}^{N}$.

Both the training and testing datasets have the same number of sensors, but they differ in their time durations, denoted as $T_{train}$ and $T_{test}$ respectively. The training dataset comprises normal data, aligning with conventional unsupervised anomaly detection methods.

The model proposed in this study utilizes historical data to predict the measured values of the next time step. The input data to predict sensor's value at time $t$, $x^{(t)} \in \mathbb{R}^{N \times w}$, is constructed using sliding window with window size $w$ over previous time series data, as follows:
\begin{equation}
    \label{eq_input_slide}
    x^{(t)}=[\boldsymbol{s}^{t-w},\boldsymbol{s}^{t-w+1},\cdots,\boldsymbol{s}^{t-1}]
\end{equation}
After the prediction, our model calculates an anomaly score for that specific time step. If the anomaly score exceeds a predefined threshold, the model outputs 1 to indicate an anomaly and 0 if the data is considered normal.

\subsection{Overview}
Our model extracts the graph structure from node embedding vectors, predicts sensor values for the next time step, and detects anomalies.
To achieve this, our model is composed of the following four main components:

\begin{enumerate}[(1)]
\item \textbf{Node Condition Embedding} uses embedding vectors to be utilized as conditions to transform source node representations and readout final node representations, and to construct the graphs structure.
\item \textbf{Graph Structure Extraction} generates graph structure based on the similarity between all pairs of the node embedding vectors.
\item \textbf{Condition-based prediction} forecasts each sensor value for the next step by employing node embedding vectors as conditions in the node conditional readout module and edge conditional node update module.
\item \textbf{Anomaly Detection} calculates anomaly scores, if the score is higher than a threshold, the model determines the time step as an anomaly.
\end{enumerate}

\begin{figure}
	\centering
		\includegraphics[width=\textwidth]{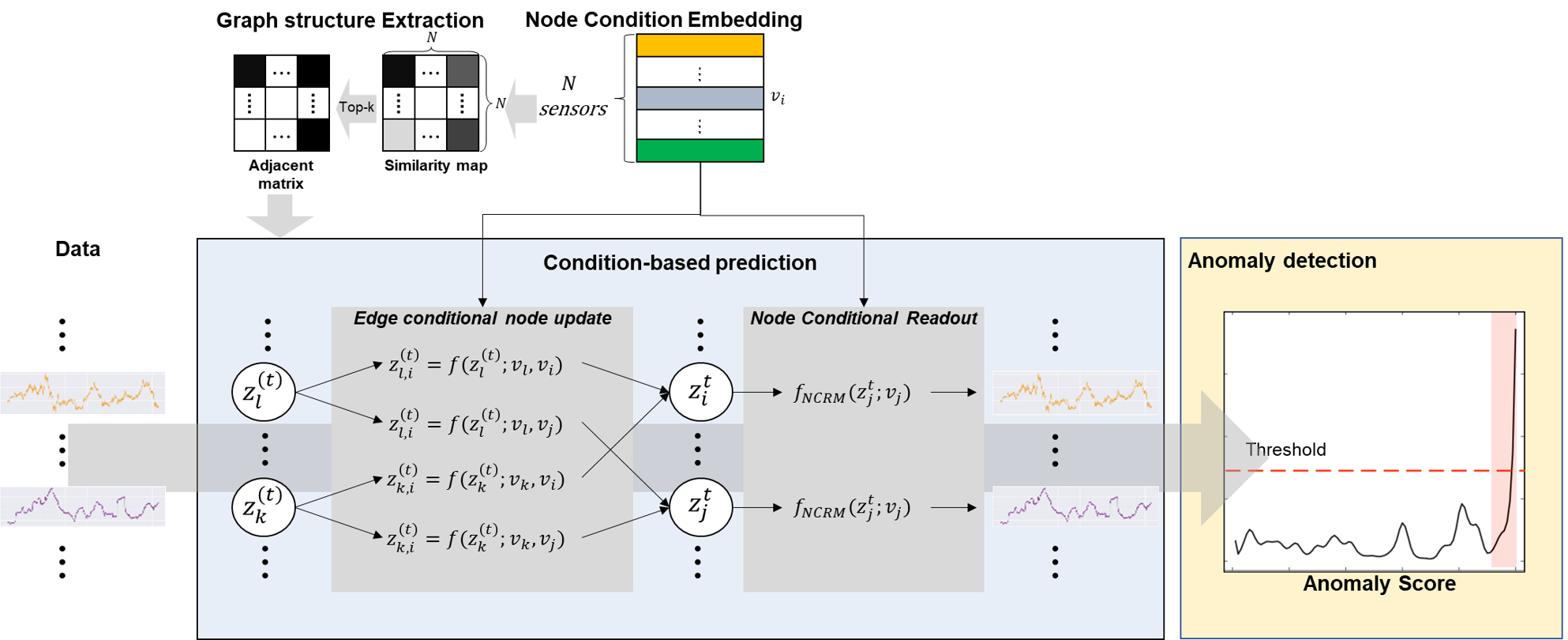}
	\caption{Overview of our model}
	\label{FIG:overview}
\end{figure}

Fig. \ref{FIG:overview} shows an overview of our model.

\subsection{Node Condition Embedding}
We introduce node embedding vectors for getting relationships between sensors, as $d_{e}$ dimensional vectors:
\begin{equation}
    \label{sensor_node_exp}
    v_{i} \in \mathbb{R}^{d_{e}}, \textnormal{for $i$}  \in \{1, 2, \cdots, N\}
\end{equation}
where $v_{i}$ represents the $i$th node embedding vector.

In \cite{deng2021graph}, the sensor-embedding vectors are trained to reflect the similar behavior of the sensors in graph attention. However, graph attention can cause node representations to become similar, which may contribute to the model's limited capability in modeling sensor behavior.

To avoid this problem, our model trains the node embedding vector without attention, instead, employing the edge conditional node update module and node conditional readout module, as described in the following section.

\subsection{Graph Structure Extraction}
To discover the relationships between sensors in systems where these relationships are not predefined, we employ the method of assessing similarities among all node embedding vectors. This similarity measurement is crucial for constructing the graph structure of the system. Specifically, for each node, we determine the Top-$k$ most similar nodes, based on the similarity score $e_{i,j}$, to form an adjacency matrix:


\begin{equation}
    \label{eq:cal_edge}
    e_{i,j}=\frac{{v}_{i} \cdot {v}_{j}}{||{v}_{i}|| \, ||{v}_{j}||}
\end{equation}
\begin{equation}
    \label{eq:adj_mat}
    A_{ij}=\mathbb{1}_{\{j \in \textit{TopK}(\{e_{i,k} : \forall k  \})\}}
\end{equation}
where \textit{TopK} is the operator used to select the indices of Top-$k$ values among the set of inputs. The value $k$ of \textit{TopK} is given by the user to adjust the sparsity level of the graph structure.

\subsection{Condition based prediction}
Condition based prediction process comprises two main modules: the edge conditional node update module (ECNUM) and the node conditional readout module (NCRM). ECNUM uses a graph-based method to generate a node representation, which is then utilized by NCRM to predict the final result. In this paper, we use MLP as ECNUM and NCRM.

A simple encoder network such as a linear or 1D CNN could be adopted to encode the input to $d_{f}$ dimensional vector, $z^{(t)} \in \mathbb{R}^{N \times d_{f}}$, from a given input $x$, as formulated in Eq. $\ref{eq:encoder}$.
\begin{equation}
    \label{eq:encoder}
    z^{(t)}=f_{\textit{enc}}(x^{(t)})
\end{equation}
Subsequently, each row of $z^{(t)}$ is adopted as the node representation.
In this paper, we utilize a linear layer as an encoder.

\subsubsection{Edge Conditional Node Update Module}
\label{ECNUM}
\begin{figure}
    \centering
    \includegraphics{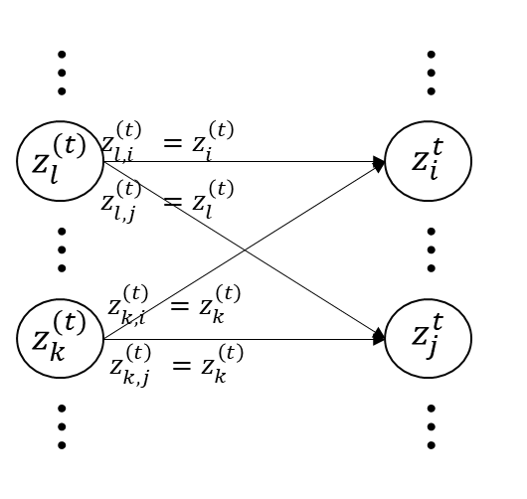}
    \caption{Method for updating target nodes in graph convolution networks: This method utilizes a uniform source node representation for different target nodes within the network. }
    \label{fig:gcn_node_update}
\end{figure}

\begin{figure}
    \centering
    \includegraphics{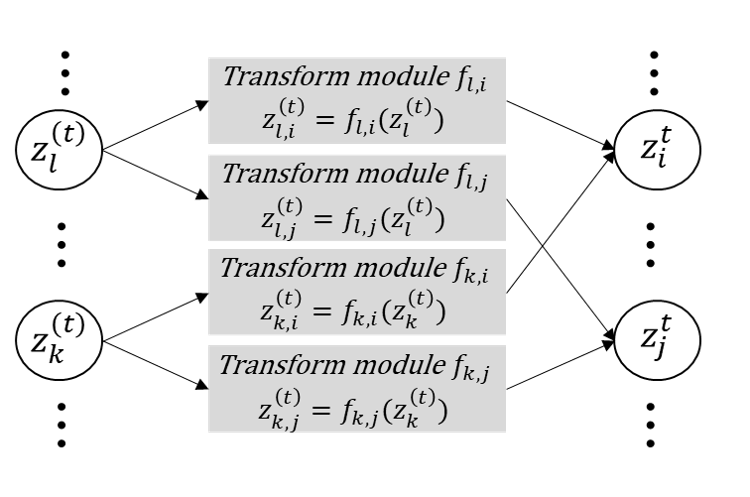}
    \caption{Naive method for updating target nodes with different source node representations: This method employs transformation modules, corresponding in number to the edges, to adapt the source node representation according to each connected edge.}
    \label{fig:trf_function_node_update}
\end{figure}

As mentioned in Section \ref{intro}, in existing models, the representation of target nodes is updated using the same representation from a source node, even when the target nodes are different, as illustrated in Fig. \ref{fig:gcn_node_update}. This means that for any given source node, its representation is uniformly applied to update the representations of all connected target nodes, regardless of their distinctiveness.
The simplest method to address this issue is to set up transformation modules for each edge, as expressed in Eq. \ref{eq:naive_idea}. 
\begin{equation}
    \label{eq:naive_idea}
    z^{(t)}_{i,j}=f_{i,j}(z_{j}^{(t)})
\end{equation}
where $z_{j}^{(t)}$ is the node representation from $x^{(t)}_{j}$, $f_{i,j}$ is the transformation function to create a new representation of source node $j$ to update target node $i$. $z_{i,j}^{(t)}$ represents the modified node representation of source node $z_{j}^{(t)}$ to update target node $i$.
With this approach, despite using the same source representation, distinct transformation modules are utilized offering diverse representations to update different target nodes, as shown in Fig. \ref{fig:trf_function_node_update}. 

\begin{figure}
\centering
    \includegraphics{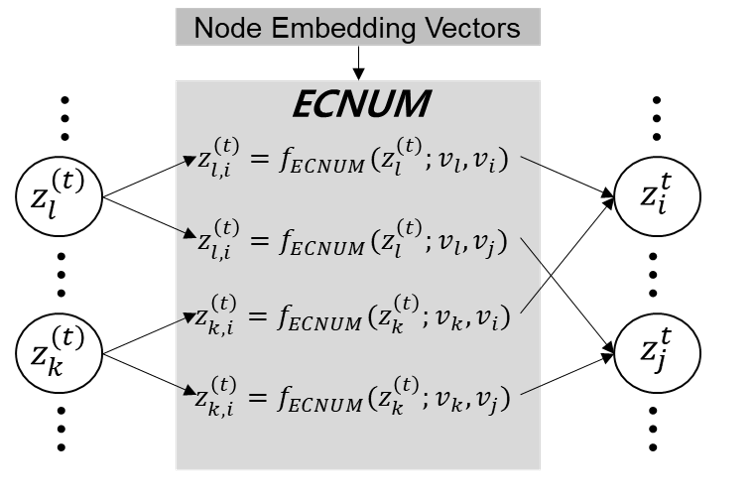}
    \caption{ECNUM Method for target node update: ECNUM utilizes a unified module approach to modify source node representations. The transformation process within this module is guided by the specific embedding vectors of both the target and source nodes.}
    \label{fig:condition_node_update}
\end{figure}

However, this method faces limitations due to the exponential increase in the number of transformation functions as the number of nodes grows, resulting in elevated computational costs and greater data requirements for training. To address these challenges, we draw inspiration from GDN \cite{deng2021graph} and conditional GAN \cite{mirza2014conditional}. GDN utilizes node embedding vectors to capture the unique characteristics of each node, while conditional GAN are capable of generating varied outputs from the same input based on specific conditions. Building on these concepts, our proposed ECNUM dynamically transforms the representation of a source node in accordance with edge conditions, as denoted by the condition embedding vectors of the source and target nodes, as follows:
\begin{equation}
    \label{eq:ecnrum}
    z_{i,j}^{(t)}=f_{\textit{ECNUM}}(z_{j}^{(t)};v_{i},v_{j})
\end{equation}
where $v_{i}, v_{j}$ are the node embedding vector of node $i,j$, respectively. $f_{\textit{ECNUM}}$ represents ECNUM.
ECNUM is a single module designed to adeptly transform source node representations to suit various target nodes, as shown in Fig. \ref{fig:condition_node_update}.

Finally, the transformed source representations, $z_{i,j}^{(t)}$ are aggregated to obtain the final node representation, as follow:
\begin{equation}
    \label{eq_gcn_update}
    z_{i}^{t}=\sigma\left(\sum_{j \in \mathcal{N}(i) \cup \{i\}} z_{i,j}^{(t)}\right)
\end{equation}
where $\mathcal{N}(i)$ is the set of neighbor node's indices of node $i$.

\subsubsection{Node Conditional Readout Module}
ECNUM enables the individual updating of each node representation with diverse source node representations, leading to a varied set of node representations.
Consider a scenario where a single readout module predicts the next values for all nodes simultaneously. In an ideal case, the readout module would effectively extract predictions from diverse node representations, given that the module has sufficient capabilities. However, if the readout module's capabilities are inadequate, it may limit the diversity of node representations to improve prediction accuracy.

\begin{figure}
\centering
    \includegraphics{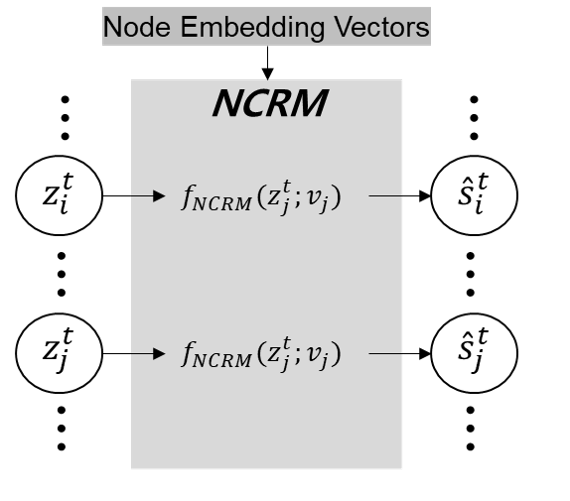}
    \caption{NCRM reads the final node representations to predict the values for the nodes at the next time step. It is a single module using node embedding vectors as conditions, enabling it to perform precise and node-specific predictions.
    }
    \label{fig:ncrm}
\end{figure}

For this reason, similar to ECNUM, we propose NCRM that takes node embedding vectors as input conditions, as illustrated in Fig. \ref{fig:ncrm}. This module can produce different results for the same node representation depending on the node embedding vector.
\begin{equation}
    \label{eq:readout}
    \hat{s}^{t}_{i}=f_{\textit{NCRM}}(z^{t}_{i};v_{i})
\end{equation}
where, $\hat{s}^{t}_{i}$ is predicted value for sensor $i$ at time $t$. $f_{\textit{readout}}$ is node conditional readout module.

To train our model, we use the Mean Squared Error as a loss function to minimize error between the predicted output $\hat{s}^{t}_{i}$ and the actual data $s^{t}_{i}$.

\begin{equation}
    \label{eq_mse_loss}
    \mathit{L}_{MSE}=\frac{1}{T_{train}-w}\sum_{t=w+1}^{T_{train}}\|s^{t}-\hat{s}^{t}\|^{2}_{2}
\end{equation}

\subsection{Anomaly Detection}
Given the values predicted from the learned graph structure, we detect anomalies using graph deviation scoring (GDS) from \cite{deng2021graph}. GDS calculates the anomaly score by comparing each sensor's scores, using the absolute error as the basis for normalization.

The first step in calculating the GDS is to calculate the absolute error between the predicted value and the actual value at time $t$ and sensor $i$:
\begin{equation}
    \label{eq_err_sensor}
    \textnormal{Err}_{i}(t)=|s_{i}^{t}-\hat{s}_{i}^{t}|
\end{equation}
The number of absolute errors at time $t$ is $N$ number of sensors. 
It is not appropriate to compare these absolute errors directly since the absolute error of each sensor exhibits a different behavior. Therefore, we normalize each absolute error as follows:
\begin{equation}
    \label{eq_norm}
    a_{i}=\frac{\textnormal{Err}_{i}(t)-\tilde{\mu}_{i}}{\tilde{\sigma}_{i}}
\end{equation}
where $\tilde{\mu}_{i}$ and $\tilde{\sigma}_{i}$ are the median and interquartile range (IQR) of the absolute error at sensor $i$. 
IQR is defined as the difference between the 1st and 3rd quartiles of set of values.
This normalization method is known to be more robust against anomalies than normalization using mean and standard variation.

To compute the anomaly score at given time $t$, we use the max function to select the most anomalous sensor and score as follow:
\begin{equation}
    \label{eq_err_fin}
    A(t)=\max_{i} a_{i}(t)
\end{equation}

The model cannot perfectly predict the sensor value; as a result, the anomaly score can exhibit a sharp change, even when the input is normal. To mitigate these changes, we use a simple moving average (SMA).

Finally, if the smoothed anomaly score is above a fixed threshold, the model labels the input as abnormal for a given time, $t$. 

Many previous methods use automatic thresholding methods, such as extreme value theory \citep{siffer2017anomaly}, to determine the threshold value. However, these methods produce different results depending on the initial set of parameters. To evaluate the anomaly detection capability of the model, we opt for a grid search method to identify an optimal threshold that maximizes the F1 score.

The pseudo-code of the edge conditional node update GNN is presented in Algorithm $\ref{alg:pseudo}$

\RestyleAlgo{ruled}
\SetKwComment{Comment}{\quad\quad\quad\quad\quad*}{}
\begin{algorithm}

    \caption{A Pseudo-code of Edge Conditional Node Update GNN}
    \label{alg:pseudo}
    \textbf{Input:} number of nodes ($N$), Top-k value ($k$), node embedding vectors ($\boldsymbol{v}$), pre-defined threshold ($\delta_{thre}$), input signal ($x^{(t)}=\{\textbf{s}^{t-w},\cdots,\textbf{s}^{t-1}\}$), ground truth signal ($\textbf{s}^{t}$)
    
    \textbf{Output:} anomaly detection result at time $t$ ($y^{t}$)
    
    
    Extract the graph structure $\textbf{A}$ following Eq. \ref{eq:cal_edge}, \ref{eq:adj_mat}\\
    
    \Comment{Encoding node representation}
    $\boldsymbol{z}^{(t)} \gets f_{enc}({x}^{(t)})$\\
    
    \Comment{Edge conditional node update}
    \For{$0 \leq i \leq $ N-1}{
    $z_{i}^{t} \gets $AGGREGATE(\{$f_{ECNUM}(z_{k}^{(t)};v_{i},v_{k})$, $\forall k \in \mathcal{N}(i)\cup \{i\}$\})
    }

    

    

    

    $\boldsymbol{z}^{t} \gets \sigma(\boldsymbol{z}^{t})$\\
    \Comment{Node conditional readout}
    $\hat{\boldsymbol{s}}^{t} \gets f_{NCRM}(\boldsymbol{z}^{t};\boldsymbol{v})$\\
    \Comment{Anomaly detection}
    \textbf{Err}$ \gets |\boldsymbol{s}^{t}-\hat{\boldsymbol{s}}^{t}|$\\
    $\tilde{\mathbf{\boldsymbol{\mu}}} \gets $ AVERAGE(\textbf{Err})\\
    $\tilde{\boldsymbol{\sigma}} \gets $ IQR(\textbf{Err})\\
    $\mathbf{a} \gets \frac{\textbf{Err}- {\tilde{\boldsymbol{\mu}}}}{\tilde{\boldsymbol{\sigma}}}$\\
    
    $A(t) \gets max(\mathbf{a})$

    $y^{t} \gets A(t) > \delta_{thre}$
    
    \Return $y^{t}$

\end{algorithm}


\section{Experiments}
\label{sec_4}
\subsection{Datasets}

Our model is evaluated using three real-world datasets. Two of these datasets, SWaT \citep{mathur2016swat} and WADI \citep{ahmed2017wadi}, are sourced from water treatment system-based test beds. These datasets are generated by simulating operator-defined attack scenarios, with evidence labels provided for each scenario. The third dataset, PSM \citep{abdulaal2021practical}, is collected from real-world server machines and is manually labeled by engineers and application experts. The PSM dataset includes both planned and unplanned anomalies

The SWaT dataset is constructed using a water treatment test bed that emulates a small-scale system. It contains 51 features, 47,515 training samples, and 44,986 test samples, with an anomaly rate of approximately 11.97\%.

The WADI dataset is built based on a water distribution test bed, which models a larger-scale system than SWaT. This dataset containes 127 features, 118,795 training samples, and 17,275 test samples, with an anomaly rate of approximately 5.99\%. 

The PSM dataset is collected internally from multiple application server nodes at eBay. It contains 25 features, 132,481 training samples, and 87,841 test samples, with an anomaly rate of approximately 27.75\%

The SWaT and WADI datasets are downsampled following the methods outlined in GDN \citep{deng2021graph}. Data are downsampled every 10s using median values to speed up the training and the labels for the test dataset are determined based on the predominant label among the down-sampled labels. The missing data is replaced by mean value. As noted by \cite{goh2017dataset}, data in the first 5–6 hours after system start-up is considered unstable; therefore, we exclude the first 2,160 data points.

\subsection{Experimental settings}
We implemented our model using PyTorch version 1.13.0, CUDA 11.6, PyTorch Geometric library 2.2.0, and PyTorch Lightning 1.9.0. A server equipped with an Intel(R) Xeon(R) Gold 6148 2.4 Hz CPU and an NVIDIA RTX A6000 GPU is used to train the model. The model is trained using Adam optimizer with a learning rate of 0.001 and beta values of 0.9 and 0.99. 
We employ an early stopping strategy with a maximum of 100 epochs and patience of 5 and select the final model with the lowest validation loss.

To identify suitable hyperparameters for the model, we employ the grid search method. The final hyperparameters are presented in Table $\ref{table:params}$.
\begin{table}[hbt!]
\centering
\caption{Hyperparameters of the model. $w$, \textit{TopK}, $d_{e}$, $d_{f}$, $n_{ECNUM}$, $n_{NCRM}$ represent the size of the sliding window, Top-$k$ factor, node embedding vector dimension, feature dimension, number of layers in ECNUM and NCRM, respectively.}
\label{table:params}
\begin{tabular}{|c|c|c|c|c|c|c|}
\hline
\textbf{Dataset} & $w$ & \textit{TopK} & $d_{e}$ & $d_{f}$ & $n_{ECNUM}$ & $n_{NCRM}$ \\ \hline
\textbf{SWaT}    & 5      & 30   & 128      & 256      & 4            & 4           \\ \hline
\textbf{WADI}    & 5      & 30   & 128      & 256      & 3            & 4           \\ \hline
\textbf{PSM}     & 3     & 25   & 128      & 128      & 1            & 2           \\ \hline
\end{tabular}
\end{table}

\subsection{Baselines}
We conduct a comprehensive comparison of our model with several baseline models. The comparison includes MAD-GAN \cite{li2019mad} from GAN-based models, USAD \cite{audibert2020usad} representing autoencoder-based models, MTAD-GAT \cite{zhao2020multivariate} and GDN \cite{deng2021graph} from GNN-based models, as well as TranAD \cite{tuli2022tranad} from transformer-based models, and GTA \cite{chen2021learning} from a hybrid model between transformer and GNN. Since the point adjustment strategy proposed by \citep{xu2018unsupervised}  has the great possibility of overestimating the model's performance, as noted in \citep{kim2022towards}, we omit this strategy when calculating the performance.

\subsection{Evaluation Metrics}
As evaluation metrics, we use metrics commonly used in previous research, including F1 score (F1), recall (Rec), and precision (Prec).

The precision measures the ratio of correctly detected anomalies to the total number of detected anomalies. This is calculated by dividing the number of correctly detected anomalies by the total number of detected anomalies.

The recall measures the ratio of correctly detected anomalies to the total number of actual anomalies in the dataset. This is calculated by dividing the number of correctly detected anomalies by the total number of actual anomalies.

The F1 score is the harmonic mean of the precision and recall, providing a balanced measure that considers both in a single metric. The precision, recall, and F1 scores can be formulated as follows:

\begin{equation}
        Prec =\frac{TP}{TP+FP}
\end{equation}
\begin{equation}
        Rec =\frac{TP}{TP+FN}
\end{equation}
\begin{equation}
        F1 =\frac{2 \times Prec \times Rec}{Prec+Rec}
\end{equation}

where $TP$, $FP$, and $FN$ denote the numbers of true positives, false positives, and false negatives, respectively.

\subsection{Results}
\begin{table}[htb!]
\caption{Average results}
\label{table:results}
\resizebox{\textwidth}{!}{%
\begin{tabular}{|c|ccc|ccc|ccc|}
\hline
\multirow{2}{*}{\textbf{Methods}} & \multicolumn{3}{c|}{\textbf{SWaT}}                                                           & \multicolumn{3}{c|}{\textbf{WADI}}                                                           & \multicolumn{3}{c|}{\textbf{PSM}}                                                            \\ \cline{2-10} 
                                  & \multicolumn{1}{c|}{\textbf{F1}} & \multicolumn{1}{c|}{\textbf{Recall}} & \textbf{Precision} & \multicolumn{1}{c|}{\textbf{F1}} & \multicolumn{1}{c|}{\textbf{Recall}} & \textbf{Precision} & \multicolumn{1}{c|}{\textbf{F1}} & \multicolumn{1}{c|}{\textbf{Recall}} & \textbf{Precision} \\ \hline
\textbf{MAD-GAN} \citep{li2019mad}                  & \multicolumn{1}{c|}{\begin{tabular}[c]{@{}c@{}}0.7206\\ (0.01219)\end{tabular}}           & \multicolumn{1}{c|}{\begin{tabular}[c]{@{}c@{}}0.5750\\ (0.01350)\end{tabular}}               & \begin{tabular}[c]{@{}c@{}}0.9658\\ (0.02368)\end{tabular}                  & \multicolumn{1}{c|}{\begin{tabular}[c]{@{}c@{}}0.2541\\ (0.00068)\end{tabular}}           & \multicolumn{1}{c|}{\begin{tabular}[c]{@{}c@{}}0.1487\\ (0.00040)\end{tabular}}               & \begin{tabular}[c]{@{}c@{}}0.8733\\ (0.00233)\end{tabular}                  & \multicolumn{1}{c|}{\begin{tabular}[c]{@{}c@{}}0.4679\\ (0.01568)\end{tabular}}           & \multicolumn{1}{c|}{\begin{tabular}[c]{@{}c@{}}0.8766\\ (0.08394)\end{tabular}}               & \begin{tabular}[c]{@{}c@{}}0.3214\\ (0.02398)\end{tabular}                  \\ \hline
\textbf{USAD} \citep{audibert2020usad}                    & \multicolumn{1}{c|}{\begin{tabular}[c]{@{}c@{}}0.7406\\ (0.00020)\end{tabular}}          & \multicolumn{1}{c|}{\begin{tabular}[c]{@{}c@{}}0.5902\\ (0.00016)\end{tabular}}              & \begin{tabular}[c]{@{}c@{}}\textbf{0.9941}\\ (0.00027)\end{tabular}                 & \multicolumn{1}{c|}{\begin{tabular}[c]{@{}c@{}}0.2543\\ (0.00076)\end{tabular}}          & \multicolumn{1}{c|}{\begin{tabular}[c]{@{}c@{}}0.1488\\ (0.00045)\end{tabular}}              & \begin{tabular}[c]{@{}c@{}}\textbf{0.8743}\\ (0.00277)\end{tabular}                 & \multicolumn{1}{c|}{\begin{tabular}[c]{@{}c@{}}0.4547\\ (0.00274)\end{tabular}}          & \multicolumn{1}{c|}{\begin{tabular}[c]{@{}c@{}}\textbf{0.9401}\\ (0.01106)\end{tabular}}              & \begin{tabular}[c]{@{}c@{}}0.2999\\ (0.00323)\end{tabular}                 \\ \hline
\textbf{MTAD-GAT} \citep{zhao2020multivariate}                & \multicolumn{1}{c|}{\begin{tabular}[c]{@{}c@{}}0.7495\\ (0.00299)\end{tabular}}          & \multicolumn{1}{c|}{\begin{tabular}[c]{@{}c@{}}0.6187\\ (0.01176)\end{tabular}}              & \begin{tabular}[c]{@{}c@{}}0.9515\\ (0.02527)\end{tabular}                 & \multicolumn{1}{c|}{\begin{tabular}[c]{@{}c@{}}0.2645\\ (0.00406)\end{tabular}}          & \multicolumn{1}{c|}{\begin{tabular}[c]{@{}c@{}}0.1857\\ (0.03330)\end{tabular}}              & \begin{tabular}[c]{@{}c@{}}0.6210\\ (0.27438)\end{tabular}                 & \multicolumn{1}{c|}{\begin{tabular}[c]{@{}c@{}}0.4716\\ (0.01595)\end{tabular}}          & \multicolumn{1}{c|}{\begin{tabular}[c]{@{}c@{}}0.6688\\ (0.12744)\end{tabular}}              & \begin{tabular}[c]{@{}c@{}}0.3834\\ (0.06902)\end{tabular}                 \\ \hline
\textbf{GDN} \citep{deng2021graph}                     & \multicolumn{1}{c|}{\begin{tabular}[c]{@{}c@{}}0.7280\\ (0.04533)\end{tabular}}          & \multicolumn{1}{c|}{\begin{tabular}[c]{@{}c@{}}0.6275\\ (0.01905)\end{tabular}}              & \begin{tabular}[c]{@{}c@{}}0.8803\\ (0.13001)\end{tabular}                 & \multicolumn{1}{c|}{\begin{tabular}[c]{@{}c@{}}0.4206\\ (0.33615)\end{tabular}}          & \multicolumn{1}{c|}{\begin{tabular}[c]{@{}c@{}}0.3110\\ (0.02991)\end{tabular}}              & \begin{tabular}[c]{@{}c@{}}0.6625\\ (0.09251)\end{tabular}                 & \multicolumn{1}{c|}{\begin{tabular}[c]{@{}c@{}}0.5476\\ (0.01774)\end{tabular}}          & \multicolumn{1}{c|}{\begin{tabular}[c]{@{}c@{}}0.7147\\ (0.06002)\end{tabular}}              & \begin{tabular}[c]{@{}c@{}}0.4479\\ (0.03584)\end{tabular}                 \\ \hline
\textbf{TranAD} \citep{tuli2022tranad}                  & \multicolumn{1}{c|}{\begin{tabular}[c]{@{}c@{}}0.5003\\ (0.23316)\end{tabular}}          & \multicolumn{1}{c|}{\begin{tabular}[c]{@{}c@{}}\textbf{0.7139}\\ (0.14616)\end{tabular}}              & \begin{tabular}[c]{@{}c@{}}0.5701\\ (0.40900)\end{tabular}                 & \multicolumn{1}{c|}{\begin{tabular}[c]{@{}c@{}}0.2524\\ (0.00078)\end{tabular}}          & \multicolumn{1}{c|}{\begin{tabular}[c]{@{}c@{}}0.1478\\ (0.00045)\end{tabular}}              & \begin{tabular}[c]{@{}c@{}}0.8680\\ (0.00266)\end{tabular}                 & \multicolumn{1}{c|}{\begin{tabular}[c]{@{}c@{}}0.4441\\ (0.01667)\end{tabular}}          & \multicolumn{1}{c|}{\begin{tabular}[c]{@{}c@{}}0.9374\\ (0.07317)\end{tabular}}              & \begin{tabular}[c]{@{}c@{}}0.2924\\ (0.02101)\end{tabular}                 \\ \hline
\textbf{GTA} \citep{chen2021learning}                     & \multicolumn{1}{c|}{\begin{tabular}[c]{@{}c@{}}0.7778\\ (0.02253)\end{tabular}}          & \multicolumn{1}{c|}{\begin{tabular}[c]{@{}c@{}}0.6411\\ (0.02949)\end{tabular}}              & \begin{tabular}[c]{@{}c@{}}0.9899\\ (0.00449)\end{tabular}                 & \multicolumn{1}{c|}{\begin{tabular}[c]{@{}c@{}}0.4083\\ (0.05022)\end{tabular}}          & \multicolumn{1}{c|}{\begin{tabular}[c]{@{}c@{}}0.3326\\ (0.06289)\end{tabular}}              & \begin{tabular}[c]{@{}c@{}}0.6195\\ (0.20943)\end{tabular}                 & \multicolumn{1}{c|}{\begin{tabular}[c]{@{}c@{}}0.5801\\ (0.00816)\end{tabular}}          & \multicolumn{1}{c|}{\begin{tabular}[c]{@{}c@{}}0.7565\\ (0.01835)\end{tabular}}              & \begin{tabular}[c]{@{}c@{}}0.4707\\ (0.01102)\end{tabular}                 \\ \hline
\textbf{Ours}                     & \multicolumn{1}{c|}{\begin{tabular}[c]{@{}c@{}}\textbf{0.8089}\\ (0.04337)\end{tabular}}          & \multicolumn{1}{c|}{\begin{tabular}[c]{@{}c@{}}0.6891\\ (0.06278)\end{tabular}}              & \begin{tabular}[c]{@{}c@{}}0.9845\\ (0.00511)\end{tabular}                 & \multicolumn{1}{c|}{\begin{tabular}[c]{@{}c@{}}\textbf{0.5730}\\ (0.04547)\end{tabular}}          & \multicolumn{1}{c|}{\begin{tabular}[c]{@{}c@{}}\textbf{0.4610}\\ (0.04782)\end{tabular}}              & \begin{tabular}[c]{@{}c@{}}0.7652\\ (0.06837)\end{tabular}                 & \multicolumn{1}{c|}{\begin{tabular}[c]{@{}c@{}}\textbf{0.6027}\\ (0.02335)\end{tabular}}          & \multicolumn{1}{c|}{\begin{tabular}[c]{@{}c@{}}0.6607\\ (0.02064)\end{tabular}}              & \begin{tabular}[c]{@{}c@{}}\textbf{0.5562}\\ (0.04222)\end{tabular}                 \\ \hline
\end{tabular}%
}
\end{table}
Table  \ref{table:results} presents a comprehensive overview of the average performance and standard deviation of both our models and the baseline models, derived from a total of 10 experiments. Our model shows performance with an average F1 score of 0.8089 on the SWaT dataset, 0.5730 on the WADI dataset, and 0.6027 on the PSM dataset. Notably, these scores represent a significant improvement over the best-performing baseline models, with increases of 4.0\% for the SWaT dataset, 36.2\% for the WADI dataset, and 3.9\% for the PSM dataset, respectively.

\begin{table}[htb!]
\caption{Experimental results of best F1 model. \dag means that value is extracted from the original paper.}
\label{table:results_best_f1}
\resizebox{\textwidth}{!}{
\begin{tabular}{|c|ccc|ccc|ccc|}
\hline
\multirow{2}{*}{\textbf{Methods}} & \multicolumn{3}{c|}{\textbf{SWaT}}                                            & \multicolumn{3}{c|}{\textbf{WADI}} &\multicolumn{3}{c|}{\textbf{PSM}}                                            \\ \cline{2-10} 
                         & \multicolumn{1}{c|}{\textbf{F1}}     & \multicolumn{1}{c|}{\textbf{Recall}} & \textbf{Precsion} & \multicolumn{1}{c|}{\textbf{F1}}     & \multicolumn{1}{c|}{\textbf{Recall}} & \textbf{Precision} & \multicolumn{1}{c|}{\textbf{F1}}     & \multicolumn{1}{c|}{\textbf{Recall}} & \textbf{Precision}\\ \hline
\textbf{MAD-GAN} \citep{li2019mad}                 & \multicolumn{1}{c|}{0.7361}   & \multicolumn{1}{c|}{0.5866} & 0.9880   & \multicolumn{1}{c|}{0.2555}   & \multicolumn{1}{c|}{0.1495} & 0.8779 & \multicolumn{1}{c|}{0.5103} & \multicolumn{1}{c|}{0.7790} & 0.3793   \\ \hline
\textbf{USAD} \citep{audibert2020usad}                    & \multicolumn{1}{c|}{0.7409} & \multicolumn{1}{c|}{0.5904} & \textbf{0.9945}   & \multicolumn{1}{c|}{0.2538} & \multicolumn{1}{c|}{0.1485} & 0.8721 & \multicolumn{1}{c|}{0.4585} & \multicolumn{1}{c|}{\textbf{0.9087}} & 0.3067   \\ \hline
\textbf{MTAD-GAT} \citep{zhao2020multivariate}                    & \multicolumn{1}{c|}{0.7566} & \multicolumn{1}{c|}{0.6261} & 0.9558   & \multicolumn{1}{c|}{0.2701} & \multicolumn{1}{c|}{0.2158} & 0.3609 & \multicolumn{1}{c|}{0.4970} & \multicolumn{1}{c|}{0.5350} & 0.4641   \\ \hline
\textbf{GDN} \citep{deng2021graph}                     & \multicolumn{1}{c|}{0.81\dag}   & \multicolumn{1}{c|}{0.6957\dag} & 0.9697\dag   & \multicolumn{1}{c|}{0.57\dag}   & \multicolumn{1}{c|}{0.4019\dag} & \textbf{0.9750}\dag  & \multicolumn{1}{c|}{0.5896} & \multicolumn{1}{c|}{0.7088} & 0.5047  \\ \hline
\textbf{TranAD} \citep{tuli2022tranad}                  & \multicolumn{1}{c|}{0.7399} & \multicolumn{1}{c|}{0.5895} & 0.9929   & \multicolumn{1}{c|}{0.2536} & \multicolumn{1}{c|}{0.1485} & 0.8721  & \multicolumn{1}{c|}{0.4812} & \multicolumn{1}{c|}{0.8496} & 0.3357  \\ \hline
\textbf{GTA} \citep{chen2021learning}                     & \multicolumn{1}{c|}{0.7948}   & \multicolumn{1}{c|}{0.6654} & 0.9870   & \multicolumn{1}{c|}{0.4694}   & \multicolumn{1}{c|}{0.3139} & 0.9215  & \multicolumn{1}{c|}{0.5929} & \multicolumn{1}{c|}{0.7501} & 0.4902  \\ \hline
\textbf{Ours}                     & \multicolumn{1}{c|}{\textbf{0.8541}} & \multicolumn{1}{c|}{\textbf{0.7593}} & 0.9761   & \multicolumn{1}{c|}{\textbf{0.6405}} & \multicolumn{1}{c|}{\textbf{0.5257}} & 0.8207  & \multicolumn{1}{c|}{\textbf{0.6285}} & \multicolumn{1}{c|}{0.6341} & \textbf{0.6230}  \\ \hline
\end{tabular}}
\end{table}
Table \ref{table:results_best_f1} shows the performance of the models with the best F1 score. Our model shows performances with an F1 score of 0.8541 on the SWaT dataset, 0.6405 on the WADI dataset, and 0.6179 on the PSM dataset.
When comparing our model to the best F1 baseline models, we observed a significant performance improvement. Specifically, our model achieved remarkable increases of 5.4\%, 12.4\%, and 6.0\% in F1 scores for the SWaT, WADI, and PSM datasets, respectively. 

These results consistently demonstrate a notably high performance improvement in the complex WADI dataset, with a 36.2\% increase in average F1 score and a 12.4\% increase in the best F1 score compared to baseline models. Moreover, the comparison with the graph-based baseline GDN model, which is similar to our approach, further validates the benefits of our model's strategy to use distinct source node representations for each target node.

\subsection{Case study}
\begin{figure}[htb!]
    \centering
    \includegraphics[width=\textwidth]{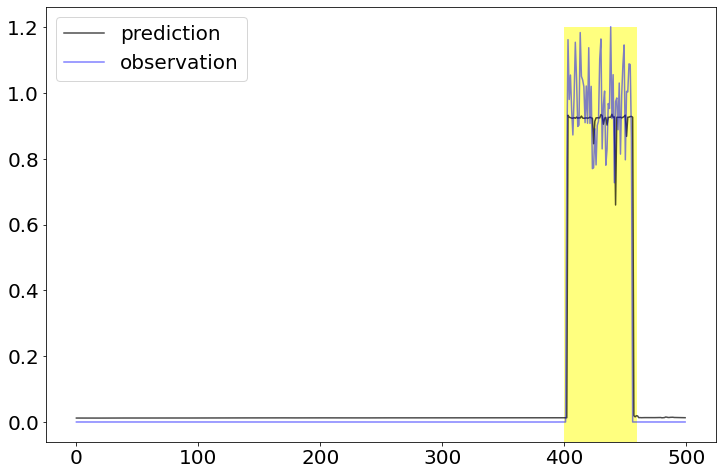}
    \caption{Predicted and observed 1\_FIT\_001\_PV sensor data of second anomaly in WADI. The yellow box indicates anomalies. Despite the sensor being under attack, detecting anomalies is challenging as there is no significant discrepancy between its predicted and observed values}
    \label{fig:fit_pv}
    
\end{figure}

To evaluate the effectiveness of our model in detecting and localizing anomalies, we analyzed the second anomaly in the WADI dataset. This anomaly is triggered by an attack that manipulates the flow sensor 1\_FIT\_001\_PV, causing it to report values different from the actual measurements. During the attack, the values remain within the normal sensor measurement range. As shown in Fig. \ref{fig:fit_pv}, it is almost infeasible to detect anomalies using the sensor independently.
\begin{figure}[htb!]
    \centering
    \includegraphics[width=\textwidth]{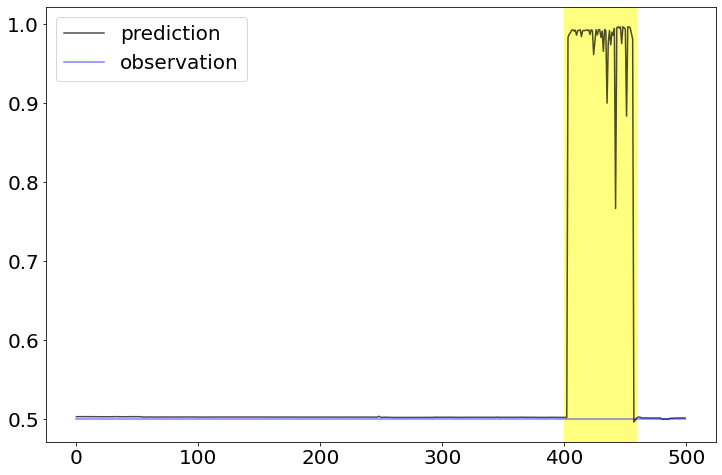}
    \caption{Predicted and observed 1\_MV\_001\_STATUS sensor data of second anomaly in WADI. The yellow box indicates anomalies. Despite not being directly attacked, a sensor can still be affected by the influence of an attacked sensor nearby. This often results in a substantial discrepancy between the predicted and observed values.}
    \label{fig:mv_status}
\end{figure}
However, during this attack period, the proposed model correctly identified 1\_MV\_001\_STATUS as a sensor with a high anomaly score, as shown in Fig. \ref{fig:mv_status}. Traditional models typically rely on attention weights to identify sensors associated with 1\_MV\_001\_STATUS. In contrast, our model uses layer-wise relevance propagation (LRP)\citep{bach2015pixel} instead of attention mechanisms to provide information about relevant sensors.

If we apply the standard LRP to our model, the edge conditional node update and the node conditional readout module do not propagate the relevance scores corresponding to the node-embedding vectors. To address this problem, we reassigned the relevance scores of the node-embedding vectors to the feature relevance scores. For the node conditional readout module, we reassigned the relevance score of the one-node embedding vector to the feature relevance score, as follows:

\begin{equation}
    \label{eq:lrp_readout}
    \hat{R}^{x_{j}}=\left(\frac{\sum_{i} R^{v_{j}}_{i}}{\sum_{i} R^{x_{j}}_{i}}+1\right) R^{x_{j}}
\end{equation}
where, $R^{x_{j}}_{i}$, $R^{v_{j}}_{i}$ are $i$th element of relevance score of feature $x_{j}$, and node embedding of node $v_{j}$, respectively.

The edge conditional node update module utilizes the summation of the relevance scores of the two node embedding vectors. The reassignment considering redundancy is conducted as follows.
\begin{equation}
    \label{eq:lrp_module}
    \hat{R}^{x_{i}}=\left(\frac{1}{|\mathcal{N}(i) \cup \{i\}|}\frac{\sum_{k} \sum_{j}( R^{S_{i,j}}_{k} + R^{T_{j,i}}_{k})}{\sum_{k} R^{x_{i}}_k}+1\right) R^{x_{i}}
\end{equation}
where $|\mathcal{N}(i) \cup \{i\}|$ is the number of edges connected to node $i$, $R^{S_{i,j}}$ is the relevance score vector for source node embedding at edge from node $i$ to node $j$, $R^{T_{j,i}}$ is the relevance score vector for target node embedding at edge from node $j$ to node $i$.

Due to the nature of the data, the inputs of some sensors may become zero. In this case, the relevance scores could also become zero, making it difficult to make accurate comparisons. To address this issue, we summed the relevance scores for each node from the relevance scores that correspond to the node representation immediately after input transformation. Since every input value is transformed independently into a node representation, the total relevance scores for each node remain consistent.

\begin{figure}[htb!]
\centering
  \includegraphics[width=\textwidth]{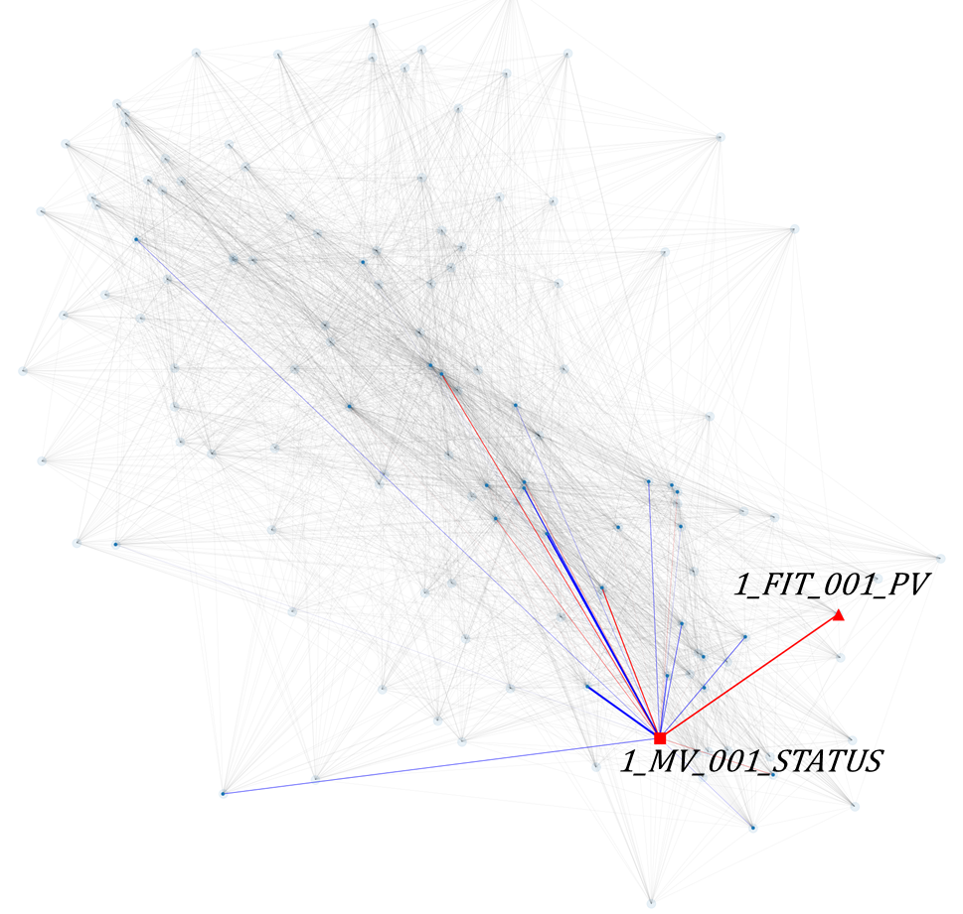}
  \caption{Learned graph structure with LRP result as edge weights for 1\_MV\_001\_STATUS node (square point) in WADI. Red edges represent positive results from LRP, while blue edges indicate negative LRP results.}
  \label{fig:sensor_infor}
\end{figure}

The graph structure learned by our model and LRP results centered on the 1\_MV\_001\_STATUS sensor are shown in Fig. \ref{fig:sensor_infor}. In the figure, the red square represents 1\_MV\_001\_STATUS and the red triangle represents 1\_FIT\_001\_PV. The red and blue edges denote the positive and negative relevance scores, respectively, and the edge thickness represents their relative magnitudes. This result confirms the high relevance of 1\_MV\_001\_STATUS and 1\_FIT\_001\_PV.

\subsection{Ablation study}
\subsubsection{Effects of node embedding condition}
An ablation study is performed to evaluate the importance of node embedding vectors as conditions by omitting their use in the edge conditional node update and readout modules. However, if all node embedding vectors are excluded, the learning of node embedding vectors fails, resulting in a fixed graph structure at its initial values. Therefore, this case is excluded from the study.
\begin{table}[htb!]
\caption{Results of ablation study for effects of node embedding vectors with SWaT and WADI}
\label{table:abl_cond}
\resizebox{\textwidth}{!}{
\begin{tabular}{|c|ccc|ccc|}
\hline
\multirow{2}{*}{\textbf{Methods}} & \multicolumn{3}{c|}{\textbf{SWaT}}                 & \multicolumn{3}{c|}{\textbf{WADI}}                 \\ \cline{2-7} 
                                  & \textbf{F1} & \textbf{Recall} & \textbf{Precision} & \textbf{F1} & \textbf{Recall} & \textbf{Precision} \\ \hline
\textbf{Original}                 & 0.8541      & 0.7593          & 0.9761             & 0.6405      & 0.5257          & 0.8207             \\ \hline
\textbf{-Target(T)}               & 0.8213      & 0.7470          & 0.9122             & 0.5716      & 0.5059          & 0.6577             \\ \hline
\textbf{-Source(S)}               & 0.7967      & 0.6676          & 0.9881             & 0.4508      & 0.3218          & 0.7541             \\ \hline
\textbf{-Readout(R)}              & 0.8497      & 0.7468          & 0.9858             & 0.5760      & 0.4238          & 0.9011             \\ \hline
\textbf{-T, R}                    & 0.8336      & 0.7240          & 0.9825             & 0.4955      & 0.4376          & 0.5689             \\ \hline
\textbf{-S, R}                    & 0.7992      & 0.7186          & 0.8999             & 0.4591      & 0.3277          & 0.7680             \\ \hline
\textbf{-S, T}                    & 0.7885      & 0.6769          & 0.9446             & 0.4676      & 0.3129          & 0.9159             \\ \hline
\end{tabular}}
\end{table}

We denote the cases where we excluded the target embedding vector in the edge conditional node update module as 'T', the cases where we excluded the source embedding vector as 'S', and the cases where we excluded the embedding vectors in the readout module as 'R'.

Ablation studies are performed on three datasets using the settings that produce the best F1 scores. In addtion, further experiments are performed on the SWaT dataset with the Top-$k$ factor set to 50, denoted as SWaT$_{50}$, to investigate the effect of a densely connected graph. The results of these studies are detailed in Table $\ref{table:abl_cond}$ and Table $\ref{table:abl_swat_50}$.

\begin{table}[htb!]
\caption{Results of ablation study for effects of node embedding vectors with PSM and SWaT$_{50}$}
\label{table:abl_swat_50}
\resizebox{\textwidth}{!}{%
\begin{tabular}{|c|ccc|ccc|}
\hline
\multirow{2}{*}{\textbf{Methods}} & \multicolumn{3}{c|}{\textbf{PSM}}                                                           & \multicolumn{3}{c|}{$\textbf{SWaT}_{50}$}                                                                          \\ \cline{2-7} 
                                  & \textbf{F1} & \textbf{Recall} & \textbf{Precision} & \textbf{F1} & \textbf{Recall} & \textbf{Precision}  \\ \hline
\textbf{Original}                 & 0.6285           & 0.6341               & 0.6230                  & 0.8114           & 0.6881               & 0.9888                                 \\ \hline
\textbf{-Target(T)}               & 0.5174          & 0.7038              & 0.4091                 & 0.7446          & 0.6161              & 0.9405                                 \\ \hline
\textbf{-Source(S)}               & 0.6110          & 0.6907              & 0.5477                 & 0.8020          & 0.6718              & 0.9943                                 \\ \hline
\textbf{-Readout(R)}              & 0.6139          & 0.6304              & 0.5982                 & 0.8037          & 0.6732              & 0.9965                                \\ \hline
\textbf{-T,R}                     & 0.5469          & 0.7119              & 0.4441                 & 0.7486          & 0.6042              & 0.9840                                 \\ \hline
\textbf{-S,R}                     & 0.6141          & 0.6748              & 0.5633                 & 0.7947          & 0.6821              & 0.9520                                 \\ \hline
\textbf{-S,T}                     & 0.5464          & 0.6694              & 0.4616                 & 0.7310          & 0.5898              & 0.9606                                 \\ \hline
\end{tabular}%
}
\end{table}
The node embedding vectors used in each module significantly contribute to performance enhancement, indicating that possessing appropriate source node representations for each target node is effective. Additionally, in the SWaT and WADI datasets, the absence of source node embedding vectors in ECNUM leads to notable performance differences. Conversely, in the PSM and SWaT$_{50}$ datasets, the absence of target node embeddings is more influential.

This variation appears to be attributable to the number of shared source nodes, which increases with larger Top-$k$ values. In less dense graph structures like SWaT or WADI, the number of overlapping source nodes is smaller when updating target nodes, allowing for effective transformation of individual source nodes to provide appropriate representations for target nodes. However, in dense graph structures like PSM or SWaT$_{50}$, where most source nodes are shared, relying solely on source node information for transformation can result in encoding similar to individual node encoding. This leads to the same issue observed in previous methods where target nodes are updated using identical source node representations regardless of the target nodes.
\subsubsection{Effects of the graph attention}
To assess the impact of graph attention, our model is compared with two models that utilize vanilla graph attention and node-wise graph attention, respectively.
The comparative models are trained by applying each graph attention method to the node representation transformed by ECNUM. The model with vanilla graph attention employs a trainable parameter $\alpha$ that is shared across all nodes to compute attention scores, as follows:
\begin{equation}
    g^{(t)}_{i}=Wz^{(t)}_{i,i}
\end{equation}
\begin{equation}
    g^{(t)}_{j}=Wz^{(t)}_{i,j}
\end{equation}
\begin{equation}
    \pi(i,j)=LeakyReLU(\alpha^{T}(g^{(t)}_{i} \oplus g^{(t)}_{j}))\\
\label{eq:graph_attn_pi}
\end{equation}
\begin{equation}
    a_{i,j}=\frac{exp(\pi(i,j))}{\sum_{k \in \mathcal{N}(i) \cup \{i\}}exp(\pi(i,k))}
\end{equation}
\begin{equation}
    z_{i}^{t}=\sigma\left(\sum_{j \in \mathcal{N}(i) \cup \{i\}} a_{i,j}z_{i,j}^{(t)}\right)
\end{equation}
where $W, \alpha$ are trainable parameters, $\oplus$ denotes concatenation.

For the node-wise graph attention model, attention scores are computed using different $\alpha_{i}$ parameters corresponding to each node. It follows the same procedure as the vanilla attention method, except for the modification of Eq. $\ref{eq:graph_attn_pi}$ as follows:

\begin{equation}
    \pi(i,j)=LeakyReLU(\alpha_{i}^{T}(g^{(t)}_{i} \oplus g^{(t)}_{j}))\\
\end{equation}

To observe the effect of graph attention on ECNUM models, we utilize the similarity of the representations transformed by ECNUM between target nodes and source nodes at time $t$, as follows:
\begin{equation}
\label{eq:total_sim}
    sim^{t}(i,j)=\frac{z_{i,i}^{(t)}\cdot z_{i,j}^{(t)}}{|z_{i,i}^{(t)}||z_{i,j}^{(t)}|}
\end{equation}

\begin{table}[]
\caption{Similarity comparison results for each dataset}
\label{table:attn_result}
\resizebox{\textwidth}{!}{
\begin{tabular}{|c|c|c|c|c|}
\hline
\textbf{Dataset}               & \textbf{Model}      & \textbf{F1} &  $sim^{t}$ Mean & $sim^{t}$ Std  \\ \hline
\multirow{3}{*}{\textbf{SWaT}} & \textbf{Origin}     & 0.8541                            & 0.5120&0.2351                                        \\
                               & \textbf{+Vanilla} & 0.8257                            & 0.8857&0.0771                            \\ 
                               & \textbf{+Node-wise} & 0.7939                            & 0.8088&0.1000                            \\ \hline
\multirow{3}{*}{\textbf{WADI}} & \textbf{Origin}     & 0.6405                            & 0.1940&0.2686                    \\
                               & \textbf{+Vanilla} & 0.5134                            & 0.6647&0.1389                    \\
                               & \textbf{+Node-wise} & 0.5678                            & 0.6361 &0.2357                            \\ \hline
\multirow{3}{*}{\textbf{PSM}} & \textbf{Origin}     & 0.6285                           & 0.3351&0.2186                    \\
                             & \textbf{+Vanilla} & 0.6114                            & 0.6532&0.1825                \\
                            & \textbf{+Node-wise} & 0.5669                           & 0.6037&0.1393                 \\
                            \hline                               
\end{tabular}}
\end{table}

The results for the mean and standard deviation of the similarity between neighboring nodes over time are shown in Table \ref{table:attn_result}. The results show that applying graph attention significantly increases the similarity between node representations encoded by ECNUM. It shows that even though ECNUM modifies the representation of the source node, graph attention acts as a constraint to ensure similarity. Vanilla graph attention, which shares the $\alpha$ parameter across all nodes to compute attention scores, shows higher similarity than node-wise graph attention, which has an individual $\alpha_{i}$ parameter at node $i$ to compute the attention score. This indicates that sharing the $\alpha$ parameter increases similarity.

\subsubsection{Efficiency of Edge Conditional Node Update Module}
To evaluate the effectiveness of employing node embedding vectors to transform source node representation dynamically, we implement a comparative model based on the GDN \citep{deng2021graph} that is most similar in structure to our model. In this comparative model, the node conditional readout module replaces the output layer of the GDN model. Furthermore, we implemented transformation modules with a structure similar to ECNUM for each edge, as described in Section \ref{ECNUM}. This implementation follows the naive idea initially described in the same section.

\begin{table}[htb!]
\centering
\caption{Results of Complexity}
\label{table:complexity}
\resizebox{\textwidth}{!}{
\begin{tabular}{c|c|c|c}
\toprule
\textbf{Model}       & \begin{tabular}[c]{@{}c@{}}\textbf{Complexity }\\ \textbf{per Layer}\end{tabular}                     & \begin{tabular}[c]{@{}c@{}}\textbf{Sequential }\\ \textbf{Operations}\end{tabular}                    & \begin{tabular}[c]{@{}c@{}}\textbf{Maximum }\\ \textbf{Path Length}\end{tabular} \\
\midrule
\textbf{Comparative Model} & $O(Ed_{f}^{2})$  & $O(E)$  & $O(E)$ \\
\textbf{ECNUM}        & $O(E(d_{f}+2d_{e})d_{f})$  & $O(1)$  & $O(1)$   \\
\bottomrule
\end{tabular}}
\end{table}

As noted in Table $\ref{table:complexity}$, the complexity of the first layer of ECNUM is $O(E(d_{f}+2d_{e})d_{f})$, compared to the comparative model's $O(Ed_{f}^{2})$. Regarding sequential operations and maximum path length, our model operates at $O(1)$, while the comparative model requires $O(E)$ for both metrics.

In terms of memory efficiency, our model requires memory proportional to $(d_{f}+2d_{e})d_{f}+(n-1)d_{f}^{2}$ since the first layer transforms input feature dimension from $(d_{f}+2d_{e})$ to $d_{f}$, and the subsequent layers operate from $d_{f}$ to $d_{f}$. In the comparative model, each transformation module, consisting of $n$ layers that transform from $d_{f}$ to $d_{f}$, is replicated across the square of the number of nodes, amounting to $N^2$ such modules. Consequently, the total memory required for all these transformation modules is proportional to $N^{2}nd_{f}^{2}$. This indicates that as the number of nodes increases, our model becomes increasingly memory-efficient compared to the comparative model, making it particularly suitable for large-scale, complex systems.

\begin{table}[htb!]
\centering
\caption{Results of efficiency}
\label{table:efficiency_result}
\resizebox{\textwidth}{!}{
\begin{tabular}{c|c|c|c}
\toprule
\textbf{Model}       & \textbf{F1}                     & \textbf{Time/Epoch}                    & \textbf{Parameters} \\
\midrule
\textbf{GDN+1 layer} & 0.8163  & 35.7236  & 43155201  \\
\textbf{GDN+2 layer} & 0.8027  & 48.0908  & 86102913  \\
\textbf{GDN+3 layer} & 0.7941  & 55.2989  & 129050625 \\
\textbf{GDN+4 layer} & 0.7736  & 66.8926  & 171998337 \\
\textbf{ECNUM}        & 0.8541  & 6.47009  & 305793   \\
\bottomrule
\end{tabular}}
\end{table}

The models are evaluated based on the F1 score, time per epoch, and the number of model parameters. 
As illustrated in Table $\ref{table:efficiency_result}$, our model surpasses the comparative model, achieving higher performance with fewer parameters and reducing time per epoch. Notably, the comparative model, equipped with four linear layers in its transformation module, possesses approximately 560 times more parameters and takes about 10 times longer per epoch. 
A trend of performance improvement is observed as the number of layers in the transformation module decreases, likely due to the reduction in the number of parameters to be learned. Additionally, we attribute the improved performance of our model to the effect of data augmentation. The transformation module learns from source node data corresponding to the edge, instead ECNUM utilizes as much source node data as there are edges.

\subsection{Sensitivity}
\begin{figure}[htb!]
    \centering
    \includegraphics[width=0.8\textwidth]{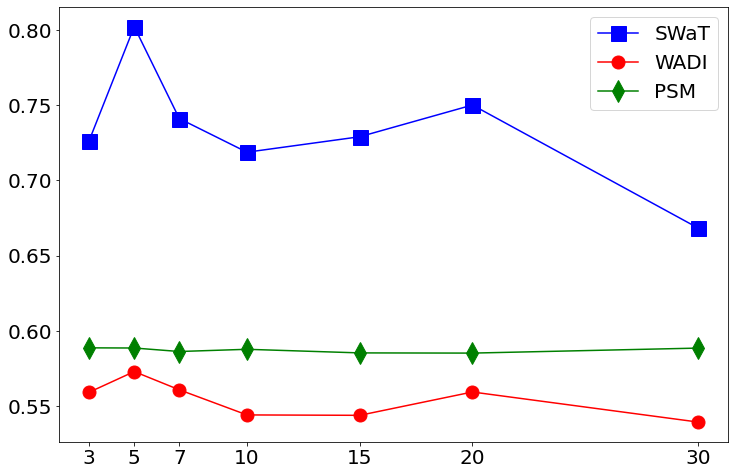}
    \caption{Sensitivity results of sliding window length}
    \label{fig:result_window}
\end{figure}
We assess the sensitivity of our model with respect to the window length and Top-$k$ factor.
We report the average F1 score and standard deviation based on 10 repetitions of the experiments. For window length, experiments are conducted with window lengths of 3, 5, 7, 10, 15, 20, and 30. Regarding Top-$k$, experiments are carried out with values of 5, 10, 15, 20, 25, 30, 35, and 40. However, for the PSM dataset, experiments are conducted only up to 25, as the dataset has a number of features lower than 30.

\begin{figure}[htb!]
    \centering
    \includegraphics[width=0.8\textwidth]{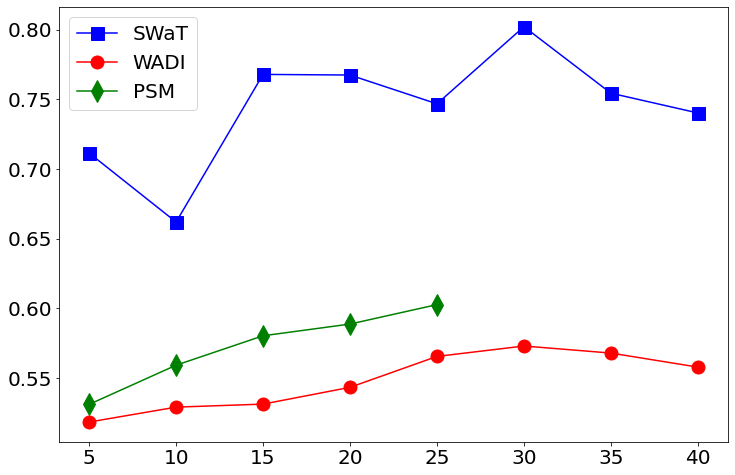}
    \caption{Sensitivity results of Top-$k$ factors}
    \label{fig:result_topk}
\end{figure}

Figure $\ref{fig:result_window}$ illustrates the results of the sensitivity analysis to the sliding window length. The figure shows that for the SWaT and WADI datasets, optimal performance is achieved with a window length of 5. In contrast, the PSM dataset exhibits stable performance across varying window lengths.

Figure $\ref{fig:result_topk}$ shows the results of the sensitivity analysis for the Top-$k$ factor. The figure shows that the performance of the models initially increases, reaches an optimal point, and then decreases. It is noteworthy that for the PSM dataset, there is no observed drop in performance, as the optimal point is identified at 25.


%
\section{Conclusion}
\label{sec_5}
In our study, we propose ECNU-GNN model that addresses the problem of updating multiple target nodes with the same source node representation and learns the graph structure using graph attention. Our model innovatively utilizes node embedding vectors to transform source node representations according to connected target nodes and learn the graph structure without using graph attention, which acts as a constraint to enforce similarity in node representations.
Our model shows a significant improvement in F1 scores across various datasets, including 5.4\% for SWaT, 12.4\% for WADI, and 6.0\% for PSM. This advancement is particularly beneficial for large, complex industrial systems, where it can aid in identifying and diagnosing system anomalies, ultimately contributing to increased operational uptime and system reliability.

Nonetheless, there are certain limitations to our method. The use of fixed node embedding vectors in our model restricts its ability to reflect dynamic graph structures that evolve over time. Additionally, applying the same Top-$k$ factor to all nodes limits our model's capability to determine the most appropriate number of neighboring nodes for each node. To overcome these challenges, our future work will concentrate on developing anomaly detection models that incorporate dynamic graph structures. These models will derive their structures from a graph inference network that learns from the temporal relationships between sensors, dynamically determining the optimal neighborhood size for each node. This advancement could enable our models to better adapt to changing data relationships, enhancing both the precision and effectiveness of anomaly detection in complex, evolving systems.

\section*{Acknowledgements}
This work was supported by the Institute for Information \& communications Technology Planning \& Evaluation(IITP) grant funded by the Korea government(MSIT) (No. 2019-0-00079, Artificial Intelligence Graduate School Program(Korea University)) and Center for Applied Research in Artificial Intelligence (CARAI) grant funded by DAPA and ADD (UD230017TD). 

\section*{Declaration of generative AI and AI-assisted technologies in the writing process}

During the preparation of this work the author(s) used ChatGPT in order to improve language and readability. After using this tool/service, the author(s) reviewed and edited the content as needed and take(s) full responsibility for the content of the publication.

 \bibliographystyle{elsarticle-num}





\end{document}